\documentclass[11pt]{article}

\usepackage[]{ACL2023}

\usepackage{times}
\usepackage{latexsym}

\usepackage[T1]{fontenc}

\usepackage[utf8]{inputenc}

\usepackage{microtype}

\usepackage{inconsolata}
\usepackage{graphicx} 

\usepackage{subcaption}
\usepackage{amsfonts} 
\usepackage{amssymb} 
\usepackage{pifont}
\usepackage{tabularx} 
\usepackage{arydshln} 
\usepackage{xcolor} 
\usepackage{mathtools} 
\usepackage{courier}
\usepackage{adjustbox} 
\usepackage{multirow} 
\usepackage{paralist} 
\usepackage{CJKutf8} 
\usepackage{booktabs} 
\usepackage{hyperref} 
\usepackage{setspace}
\usepackage{enumerate}
\usepackage{bbding} 
\usepackage{wasysym}
\usepackage{multirow}
\usepackage{makecell}
\usepackage[most]{tcolorbox}
\usepackage{xcolor} 

\definecolor{calloutcolor}{HTML}{D6EDFF} 
\tcbset{
  callout/.style={
    enhanced,
    colback=calloutcolor!40, 
    colframe=calloutcolor!40, 
    boxrule=0mm, 
    colbacktitle=calloutcolor, 
    coltitle=black, 
    arc=2mm, 
    width=\linewidth, 
    boxsep=2mm, 
    left=1mm, right=1mm, top=1mm, bottom=1mm, 
    fonttitle=\bfseries, 
    title={#1}, 
    toptitle=0mm, 
    bottomtitle=0mm, 
  }
}

\title{Breaking Language Barriers:\\ Cross-Lingual Continual Pre-Training at Scale}

\author{Wenzhen Zheng$^{1,\dagger}$, Wenbo Pan$^{2,\dagger}$, Xu Xu$^{3,\dagger}$, Libo Qin$^4$, Li Yue$^5$, Ming Zhou$^5$ \\
  $^1$Chinese Academy of Sciences
  $^2$Harbin Institute of Technology 
  $^3$Peking University \\
  $^4$School of Computer Science and
  Engineering,
  Central South University \\
  $^5$Langboat Inc. \\
  \texttt{zhengwenzhen@amss.ac.cn}, 
  \texttt{wenbopan@proton.me}, \\
  \texttt{xuxu\_pkuse@stu.pku.edu.cn}, 
  \texttt{lbqin@csu.edu.cn}, \\
  \texttt{\{yueli, zhouming\}@langboat.com} 
}

\begin{document}
\maketitle
\begin{abstract}
In recent years, Large Language Models (LLMs) have made significant strides towards Artificial General Intelligence. However, training these models from scratch requires substantial computational resources and vast amounts of text data. In this paper, we explore an alternative approach to constructing an LLM for a new language by \textit{continually pre-training} (CPT) from existing pre-trained LLMs, instead of using randomly initialized parameters. Based on parallel experiments on 40 model sizes ranging from 40M to 5B parameters, we find that 1) CPT converges faster and saves significant resources in \textit{a scalable manner}; 2) CPT adheres to an  extended scaling law derived from \citet{hoffmann2022training} with a joint data-parameter scaling term; 3) The compute-optimal data-parameter allocation for CPT markedly differs based on our estimated scaling factors; 4) The effectiveness of transfer at scale is influenced by training duration and linguistic properties, while robust to \textit{data replaying}, a method that effectively mitigates catastrophic forgetting in CPT. We hope our findings provide deeper insights into the transferability of LLMs at scale for the research community.
\end{abstract}

\section{Introduction}

\footnotetext[1]{$\dagger$Work done during internship at Langboat Inc. Authors contributed equally.}

\begin{figure*}[htbp]
\centering
\includegraphics[width=\textwidth]{"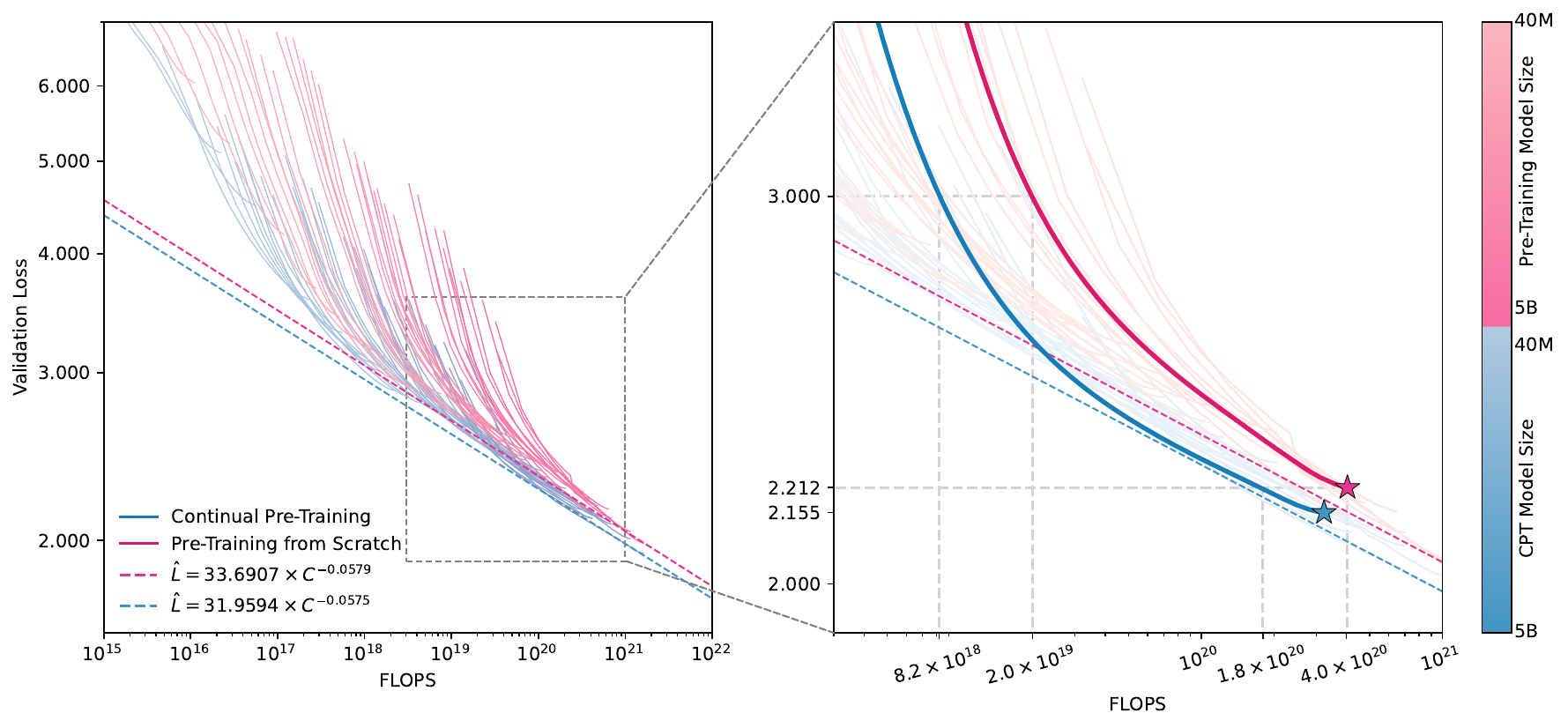"}
\caption{Loss curves of \textit{pre-training} and \textit{continual pre-training} (CPT) across different model sizes. All models are pre-trained on Chinese text while CPT models are initialized from pre-trained English checkpoints. Dashed lines predict optimal loss at each computation level, as estimated in Section~\ref{sec:lc-re}. \textit{(Left)} Overlapped loss-compute power-law visualization, with each line representing one model. \textit{(Right)} CPT LLM (2B parameters) reaches the same loss with approximately 50\% fewer FLOPs.}
\label{fig:scaling}
\end{figure*}

In recent years, Large Language Models (LLMs) pre-trained on web-scale corpora have achieved significant success in various language tasks~\cite{radford2019language,brown2020language,achiam2023gpt}. As the scale of pre-training increases, LLMs have exhibited remarkable abilities, particularly in transferring knowledge across different domains~\cite{wei2022emergent, tan2018survey}.

Training an LLM from scratch is prohibitively expensive. To address this, some practitioners leverage \textit{transfer learning} to adapt LLMs to new domains or tasks. This usually involves fine-tuning the models on a small dataset within the target domain. Previous works have showcased multiple benefits of transfer learning in fine-tuning when the transfer gap is small, including faster convergence and better final performance~\cite{zhang2024scaling, hernandez2021scaling}.
However, it remains unclear if these benefits hold when fine-tuning on massive data or across large distribution shifts (e.g., different languages). This becomes a crucial consideration if one aims to efficiently build an LLM using transfer learning, especially when there is a sufficient amount of data available from different distributions.

To fill this gap, we investigate training LLMs with transfer learning on large pre-training corpora. To be specific, we create LLMs for a new language by using pre-trained LLMs as initialization instead of starting from scratch. We refer to this approach as \textit{continual pre-training} (CPT). The motivation for our work stems from the inherent ability of meta-knowledge to transfer across various languages \cite{pan2009survey,zhuang2020comprehensive,tang2020multilingual,eronen2023zero}. By leveraging this transferability, LLMs can use existing linguistic knowledge to enable more efficient training. 

In this paper, we conduct pre-training with parameter sizes ranging from 40M to 5B, spanning 40 different sizes, to systematically study the effect of CPT at different conditions and scales. Specifically, we use English as the \textit{source language} for the source model and Chinese as the \textit{target language} for CPT. We compare two different training strategies:

\begin{enumerate}

\item \textbf{Training from Scratch}: The pre-training of Chinese LLM begins with completely randomly initialized parameters and is trained using Chinese language corpora.

\item \textbf{Continual Pre-Training (CPT)}: The parameters of a Chinese LLM are initialized with those from an equivalent English LLM and then trained using Chinese language corpora.

\end{enumerate}
Figure~\ref{fig:scaling} summarizes our main training results. We find that, CPT models of different sizes exhibit a power-law relationship between loss and compute similar to models trained from scratch, but achieve lower loss at each computational level. For models of a given parameter size, CPT consistently outperforms training from scratch, particularly during the initial stages. Throughout the whole training process, CPT saves 25\% to 50\% of tokens when achieving the same loss.

Our main focus lies in the comparative analysis between the two strategies, including their scaling behaviors, the robustness of scaling, and their corresponding impact factors. We also study the technique of \textit{data replaying}~\cite{Ibrahim2024SimpleAS} to mitigate catastrophic forgetting in CPT. Data replaying involves replaying a portion of the source language data during the training of the target language model. Finally, to explain and model the scalings under different settings, we fit a new extended scaling law for CPT, derived from \citet{hoffmann2022training}. Our findings are outlined as follows:

\begin{itemize}
  \item CPT demonstrates persistent training advantages even at the pre-training scale. For example, after training on 70B tokens, the 5.5B model with CPT reaches the same loss as a model trained from scratch with 110B tokens.
  \item Our extended scaling law more accurately captures the scaling behavior in CPT, revealing a positive multiplicative joint scaling effect between data and parameter size.
  \item Based on the extended scaling law, we determine the compute-optimal data-parameter allocation for CPT, which favor larger parameter sizes over larger datasets compared to training from scratch.
  \item The transfer scaling effect in CPT is stronger with fewer training tokens or when the target language is more similar to the source language, but robust to \textit{data replaying}.
  \item CPT is susceptible to catastrophic forgetting; however, replaying 10\% to 30\% of the source language data effectively mitigates this issue.
  \end{itemize}

\section{Setup}
\label{sec:setup}
\subsection{Training Framework}

\begin{table*}[ht]
\centering
\caption{Training configurations for pre-training. All three sets of models are trained with identical parameter sizes, which cover 40 sizes spanning from 50M to 5.5B. Note that the batch size is based on token counts.}
\begin{tabular}{llll}
\toprule
Model Set & Initialization & \makecell{Training \\ Language} & \makecell{Parameter Size \& \\ Batch Size \textsubscript{(Same for Each Set)}} \\
\midrule
Source Checkpoints & Random & English & 50M-1B\textsuperscript{(23 models)}
,1M \\
Pre-trained from Scratch  & Random & Chinese & 1B-2.5B\textsuperscript{(12 models)}, 2M \\
Continually Pre-trained & Source Checkpoints & Chinese & 2.7B-5.5B\textsuperscript{(5 models)}, 4M \\
\bottomrule
\end{tabular}
\label{tab:model_settings}
\end{table*}

To compare the transfer effects in CPT versus pre-training from scratch, we train two sets of models with the same parameter sizes. Additionally, another set of model checkpoints is trained in the source language to serve as the initialization for the continually pre-trained models. The training configurations for the three sets of models are shown in Table~\ref{tab:model_settings}.

To simplify the experiments, we use identical training strategies for all three pre-training sets, including learning rate schedules, batch sizes and trained token counts. All models are pre-trained with a context length of 2048 and undergo training on tokens equivalent to 20 times the model size (e.g., a 5B model is trained on 100B tokens). Although this is far from the extensive pre-training seen in recent practices ~\cite{touvron2023llama2}, as outlined in \citet{hoffmann2022training}, the 20x trained token count is sufficient to demonstrate the loss-data scaling relationship. Our learning rate (LR) schedule features a cosine LR decay from a maximum LR of $2 \times 10^{-4}$ and an LR warm-up, which increases the LR to the maximum in the first 5\% of the training session. We use different batch sizes for different parameter sizes, as shown in Table~\ref{tab:model_settings}.

\subsection{Model and Data}

\paragraph{Model Architecture} We adopt the same decoder-only Transformer architecture as LLaMA2~\cite{touvron2023llama2} for all pre-training. We choose LLaMA2 because it is widely studied and proven to scale well across different parameter sizes. Following \citet{muennighoff2023scaling}, we derive architectural parameters for models of each parameter size, which are listed in Appendix~\ref{app:struct}. 

\paragraph{Tokenizer} The tokenizer from LLaMA2 doesn't properly represent common Chinese characters and tends to over-slice Chinese sentences. To prevent this, our tokenizer is trained on the bilingual pre-training corpus we used for the CPT experiments. The tokenizer is trained with SentencePiece~\cite{kudo2018sentencepiece} and has a vocabulary size of 36,152, including common Chinese word pieces. 

\paragraph{Data Sources} Our English training data is primarily sampled from the RedPajama dataset \cite{together2023redpajama}, while the Chinese training data was sampled from Common Crawl~\cite{commoncrawl} and WuDao~\cite{wudao2022}. The raw text undergoes filtering and deduplication processes, which is similar to RedPajama~\cite{together2023redpajama}. To study langauge robustness of the CPT strategy, we also conduct experiments on other languages, including French and Russian. We take their corresponding subsets from mC4 \cite{2019t5} as pre-training data. An total of $10^6$ tokens are held out from each respective training set as validation sets, remaining consistent across different models.

\subsection{Evaluation Tasks}
Throughout experiments, we primarily use \textit{cross-entropy loss} on held-out validation sets as an indicator of model performance. To further validate the generalizability of CPT, we also evaluate LLMs using widely adopted language modeling benchmarks. To assess models in different languages, we choose multilingual versions of existing benchmarks, including XNLI~\citep{conneau2018xnli}, Multilingual Winograde~\cite{Sakaguchi2019WinoGrande}, Multilingual Hellaswag~\cite{dac2023okapi}, XStorycloze~\cite{Lin2021FewshotLWStorycloze}, XCopa~\cite{Ponti2020XCOPAAM}, and PiQA~\cite{Bisk2019PIQARA}. Note that for French and Russian, we exclude XCopa~\cite{Ponti2020XCOPAAM} and PiQA~\cite{Bisk2019PIQARA} as they do not contain splits for these two languages. All evaluations are performed under zero-shot settings. We report normalized accuracy as the metric for each task.

\section{Methodology}
\label{sec:method}

\subsection{Scaling Law for Pre-Training from Scratch}

We follow the Chinchilla Scaling Law~\cite{hoffmann2022training} to express cross-entropy loss ($L$) as a function of parameters ($N$) and training tokens ($D$):

\begin{equation}
\begin{aligned}
L(N, D) = & \ E + \frac{A}{N^\alpha} + \frac{B}{D^\beta}
\label{eq:chinchilla_2}
\end{aligned}
\end{equation}

\noindent where $\{E, A, B, \alpha, \beta\}$ are learned variables. The Chinchilla law further determines the optimal allocation of compute (C) to $N$ and $D$ as:

\begin{equation}
\begin{aligned}
    N_{\text{opt}}(C) &= G\left(\frac{C}{6}\right)^a\\
    D_{\text{opt}}(C) &= G^{-1}\left(\frac{C}{6}\right)^b
    \label{eq:optimal}
\end{aligned}
\end{equation}

\noindent where $ G = \left(\frac{\alpha A}{\beta B}\right)^{\frac{1}{\alpha+\beta}}$, with $a = \frac{\beta}{\alpha + \beta}$, $ b = \frac{\alpha}{\alpha + \beta}$. The ratio of $a$ to $b$ represents the optimal data-to-parameter size allocation.

Additionally, as shown in \citet{kaplan2020scaling}, the optimal loss, independent of parameters and data, also scales with compute $C$ following a power-law relationship:

\begin{equation}
\begin{aligned}
L_{opt}(C) = & \ E' + \frac{A'}{C^\gamma}
\label{eq:lc}
\end{aligned}
\end{equation}

\subsection{Scaling Law for Continual Pre-Training}
\label{sec:modification}

\begin{table*}[t]
\centering
\small
\setlength{\tabcolsep}{4pt}
\caption{\label{tab:fit_results} Comparison of parameter estimation and optimization coefficients for Equation~\ref{eq:oursloss} and Equation~\ref{eq:fit}. For Continual Pre-Training, parameters $E$, $A$, and $\alpha$ are fixed based on values from Training from Scratch.}
\subcaptionbox{\label{tab:simplified_fit_results} Estimations for Equation~\ref{eq:oursloss}.}{
\begin{tabularx}{0.85\textwidth}{lXXXXXX}
\toprule
Model & $E$ & $A$ & $B$ & $\alpha$ & $\beta$ & $\gamma$ \\
\midrule
Training from Scratch & 1.55 & 420.0 & 719.5 & 0.40 & 0.30 & - \\
Continual Pre-training & 1.55 & 420.0 & 433.3 & 0.40 & 0.20 & 0.08 \\
\bottomrule
\end{tabularx}}

\vspace{\baselineskip}

\subcaptionbox{\label{tab:coefficients} Approximated optimization coefficients for Equation~\ref{eq:optimal}.}{
\begin{tabularx}{0.85\textwidth}{Xcc}
\toprule
Model & $\text{Coeff. } a \text{ where } N_{\text{opt}} \propto C^{a}$ & $\text{Coeff. } b \text{ where } D_{\text{opt}} \propto C^{b}$ \\
\midrule
Training from Scratch & 0.429 & 0.571 \\
Continual Pre-training & 0.385 & 0.615 \\
\bottomrule
\end{tabularx}}
\end{table*}

The Chinchilla law assumes that LLM pre-training is initialized with no prior knowledge, which does not apply to continual pre-training (CPT). To extend the Chinchilla law for CPT, we incorporate insights from \citet{hernandez2021scaling}, introducing an \textit{effectively transferred data} term. According to \citet{hernandez2021scaling}, effective data transfer is modeled as $k(D_F)^\alpha (N)^\beta$, capturing the idea that larger models store more transferable knowledge. Thus, we extend the $D$ term to include a multiplicative joint effect of both $D$ and $N$, resulting in our CPT loss function:

\begin{equation}
\begin{aligned}
L(N, D) = E + \frac{A}{N^\alpha} + \frac{B'}{D^{{\beta'}} N^\gamma}
\label{eq:oursloss}
\end{aligned}
\end{equation}

Accordingly, we update Equation~\ref{eq:optimal} for the extended scaling law:

\begin{equation}
\begin{aligned}
G = \left(\frac{\alpha A}{({{\beta'}} - \gamma) B'}\right)^{\frac{1}{\alpha+{{\beta'}}-\gamma}}, \\ a = \frac{{{\beta'}}}{\alpha + {{\beta'}} - \gamma}, b = \frac{\alpha - \gamma}{\alpha + {{\beta'}} - \gamma}
\label{eq:fit}
\end{aligned}
\end{equation}

Note that we do not update $A$, $E$, and $\alpha$ during optimization for CPT. Preliminary experiments show minimal impact of CPT on the $N$ term, so we keep these variables from Equation~\ref{eq:chinchilla_2} to reduce variance. Empirical experiments demonstrate that the extended scaling law achieves a lower fitting error than the Chinchilla law for CPT. Additionally, the introduced data-parameter joint term captures meaningful features in scaling behavior, as shown in Section~\ref{sec:measure}. We provide fitting error comparison for both scaling laws in Appendix~\ref{sec:fiterror}, where we show that extended scaling law performs better for CPT. We also give more theoretical analysis and interpretation of the extended scaling law in Appendix~\ref{app:explan}.

\subsection{Parametric Fit}
\label{sec:fit}
To fit the learnable variables in Equation~\ref{eq:oursloss}, we minimize the Huber loss~\cite{huber1992robust} between predicted and observed log loss, with $\delta$ set to $10^{-3}$. For pre-training from scratch, we minimize Equation~\ref{eq:chinchilla_2}:
\begin{equation}
\begin{aligned}
\min_{a,b,e,\alpha,\beta} \sum_{\text{Run } I} & \text{Huber}_\delta \left( \text{LSE}(a - \alpha \log N_i, \right. \\
& \left. b - \beta \log D_i, e) - \log L_i \right)
\end{aligned}
\end{equation}

\begin{figure}[tb]
    \centering
    \includegraphics[width=\linewidth]{"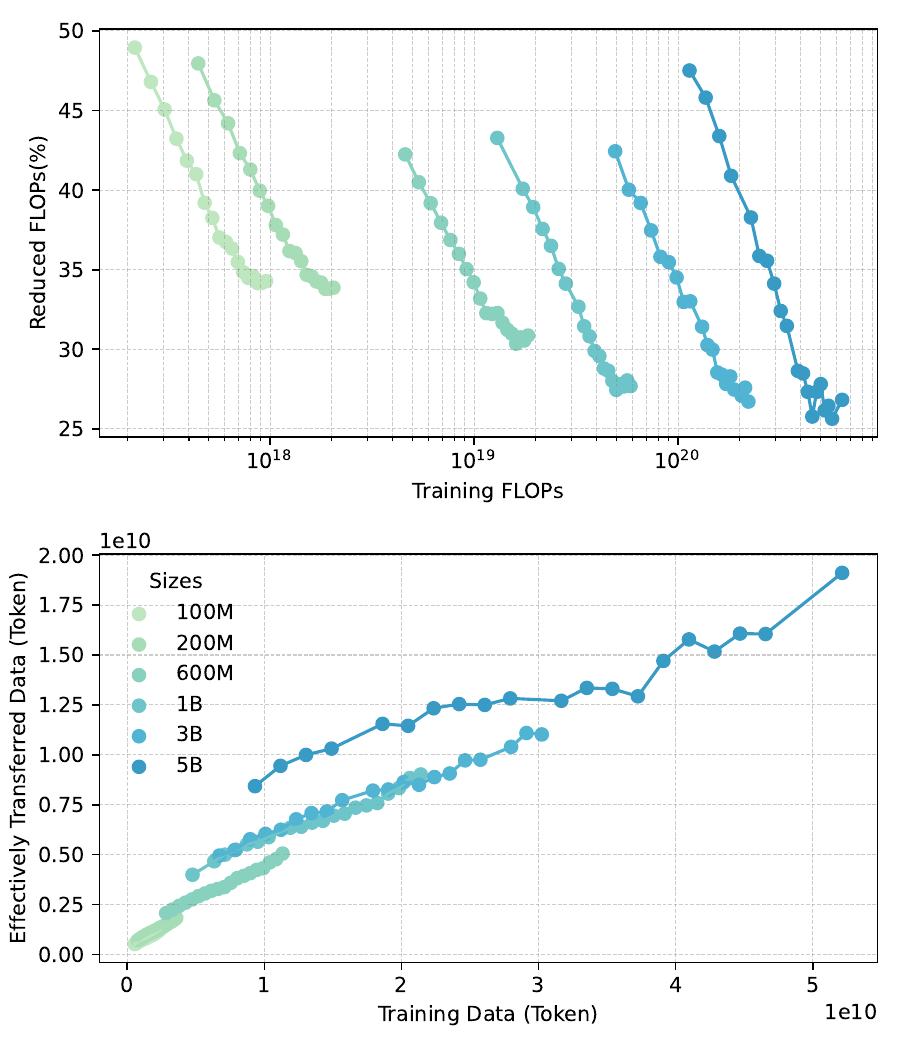"}
    \caption{Reduced computational resources (top) and data consumption (bottom) with CPT. Only a subset of models of typical sizes is displayed for simplicity. \textit{(Top)} Percentage reduction in FLOPs $C$ relative to pre-training from scratch $PT$, as estimated by $(C_{PT} - C_{CPT}) / C_{PT}$ at the same loss level for both strategies. \textit{(Bottom)} Effectively Transferred Data, calculated by subtracting the tokens $D$ used by CPT from those used in pre-training from scratch at the same loss level, i.e. $D_{PT} - D_{CPT}$. }
    \label{fig:saved}
\end{figure}

\noindent where $LSE$ is the \textit{log-sum-exp} operator. We set $A = \exp({a})$, $B = \exp({B})$, $B' = \exp({b'})$, and $E = \exp({e})$. For continual pre-training, using the fixed values of $a$, $\alpha$, and $e$ from the previous optimization step, we subsequently optimize ${B', \beta', \text{and } \gamma}$ in Equation~\ref{eq:oursloss}:

\begin{equation}
\begin{aligned}
\min_{b',{{\beta'}},\gamma} \sum_{\text{Run } I} & \text{Huber}_\delta \left( \text{LSE}(a - \alpha \log N_i, \right. \\
& \left. b' - {{\beta'}} \log D_i - \gamma \log N_i, e) - \log L_i \right)
\end{aligned}
\end{equation}

We use the Optuna library for hyperparameter search and the L-BFGS algorithm~\cite{nocedal1980updating} for optimal local search, yielding the best hyperparameters. The final parameter values are presented in Table~\ref{tab:simplified_fit_results}, and the optimized allocation coefficients are shown in Table~\ref{tab:coefficients}.


\section{Results}
\label{sec:benefits}

\subsection{CPT Reaches Lower Loss Throughout Training}

Figure~\ref{fig:scaling} reports the validation loss over training for all trained models. It can be seen that pre-training language models from existing checkpoints generally yield lower loss given certain compute constraints. This effect exists across both various model sizes and training stages of the same model. At the start of training, CPT converges significantly faster, advancing pre-training from scratch by orders of magnitudes. The absolute difference of loss becomes smaller as training continues, but a substantial gap in loss persists. Note that Figure~\ref{fig:scaling} is presented on a logarithmic scale. This gap may require several orders of magnitude more iterations before it disappears.

\subsection{CPT Preserves Loss-Compute Scaling Relationship}
\label{sec:lc-re}
As indicated by Equation~\ref{eq:lc}, optimal validation loss scales with compute following a power-law relationship. We conducted parametric fits for CPT and pre-training from scratch on Equation~\ref{eq:lc}, using the lowest loss at each compute level. The fit results are depicted as dotted lines in Figure~\ref{fig:scaling}. For pre-training from scratch, the relationship is represented by \(L = 33.69907 \times C^{-0.0579}\). In comparison, the loss for CPT is lower, described by \(L = 31.9594 \times C^{-0.0575}\).

The results of the parametric fit indicate that the advantage of lower loss is consistent across each unit of compute expended. This is supported by the significantly reduced coefficient term (from 33.69907 to 31.9594) and the nearly unchanged exponent (from -0.0579 to -0.0575). The nearly unchanged exponent suggests that CPT does not alter the underlying dynamics of the loss-compute relationship, but rather provides an advantageous initial condition.

\begin{figure*}[h]
    \centering
    \includegraphics[width=\textwidth]{"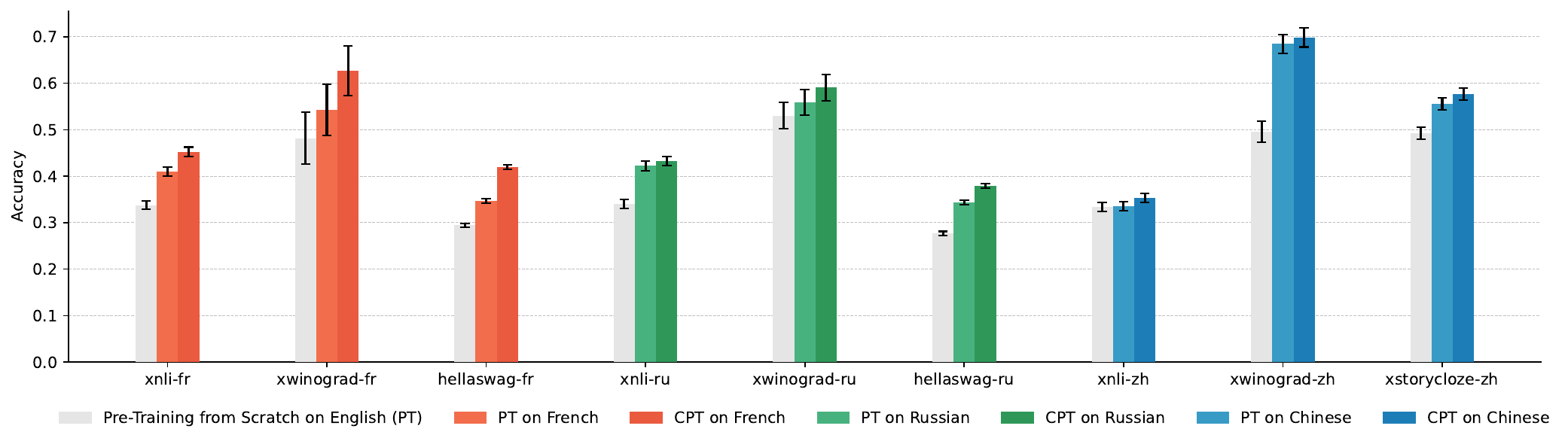"}
    \caption{Zero-shot evaluation for pre-trained and continually pre-trained (CPT) models of different languages. CPT models of various languages are initialized from the same checkpoint (light gray).}
    \label{fig:macc}
\end{figure*}

\subsection{Extended Scaling Law Measures Effectively Transferred Data in CPT}
\label{sec:measure}
    
We conducted a further analysis to study the impact of individual factors, specifically data and model size, on loss. Table~\ref{tab:simplified_fit_results} compares the estimated parameters for CPT with those for training from scratch. As discussed in Section~\ref{sec:modification}, only the parameters in the term $\frac{B'}{D^{\beta'} N^\gamma}$ are updated for CPT. For CPT, the parameters are $B = 433$, $\gamma = 0.08$, and $\beta = 0.20$. The lower $\beta$ and positive $\gamma$ suggest that in CPT, the cross-lingual transfer effect positively correlates with parameter size.

In Figure~\ref{fig:saved}, we measure the transferred training FLOPs and data during CPT to visualize the scaling transfer effect of parameter size, which corroborates our theoretical results. We find that the percentage of reduced training FLOPs steadily decreases during the individual training process, resulting in 25\% to 50\% FLOPs saved during CPT. On the other hand, effectively transferred data linearly increases with training tokens, with larger models reducing more training FLOPs and data during CPT, indicating a stronger transfer effect. A plausible explanation could be that a larger optimization space contains more linguistic-agnostic knowledge that can transfer more easily.

\subsection{CPT Models Generalize to Downstream Tasks}

Besides validation losses, we also evaluate cross-lingual CPT on several multi-lingual benchmarks. Using 1.4B parameters, we continually trained models in French (Fr.), Russian (Ru.), and Chinese (Zh) from the same English checkpoint and compared them to models trained from scratch and the original English checkpoints. The results, shown in Figure~\ref{fig:macc}, reveal that CPT improves performance across all languages.

Our results showed that in all three languages tested, the models enhanced through CPT consistently outperformed those trained from scratch, demonstrating improved performance across various languages and benchmarks.

We find that French models benefit the most from CPT. This is likely due to the high similarity between French and English, which share many common words and grammatical structures, facilitating more effective cross-lingual transfer compared to Russian and Chinese.

\begin{tcolorbox}[callout={Key Takeaways}]
    \begin{enumerate}[\textbullet]
        \item Continual pre-training converges to lower loss faster throughout training, saving 25\% to 50\% of training FLOPs.
        \item The transfer effect is most pronounced in the early stages and positively correlated with parameter size.
        \item The effect generalizes well to downstream evaluations, with languages more similar to English experiencing greater benefits.
    \end{enumerate}
\end{tcolorbox}

\section{Discussion}
\label{sec:diss}

\subsection{What is the Compute-Optimal Allocation between Parameter Size and Data?}
\label{sec:compute}

\begin{figure*}[htb]
    \centering
    \includegraphics[width=\textwidth]{"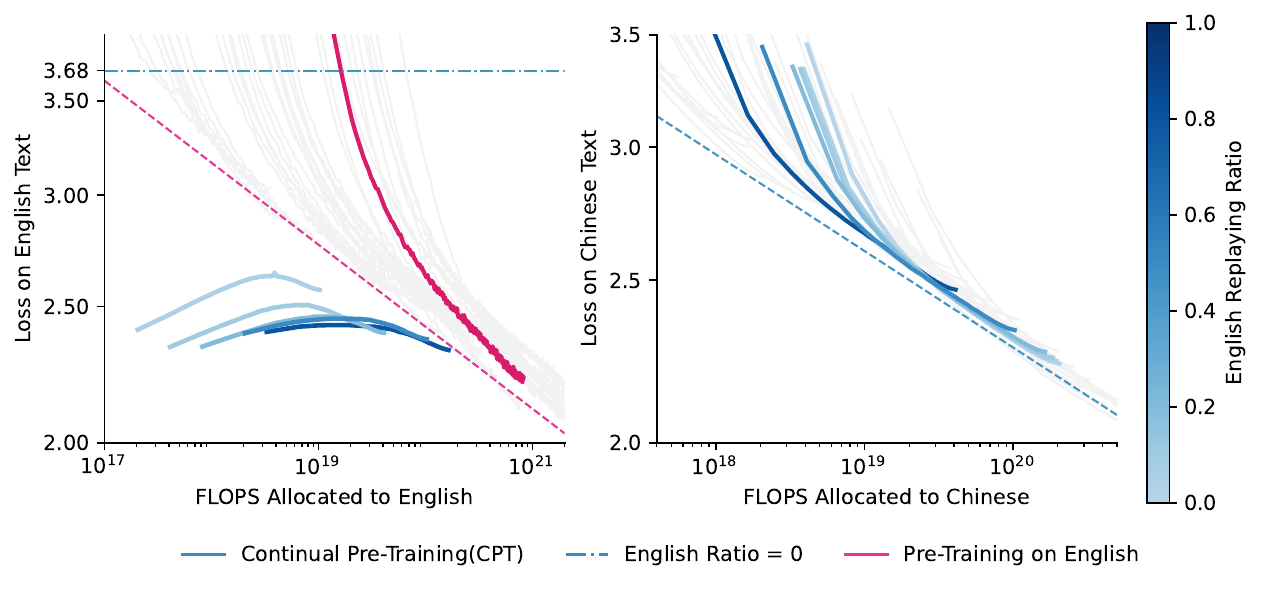"}
    \caption{Scaling of CPT with different English replaying ratios. Each blue line represents a 1.4B model continually pre-trained with various replaying ratios and evaluated on two validation sets: English (\textit{left}) and Chinese (\textit{right}). Models with English replaying ratios of 1\%, 5\%, 10\%, 20\%, 50\%, and 80\% are shown from light to dark blue, respectively. FLOPs allocated to each language are calculated by multiplying the corresponding language ratios by the total FLOPs.}
    \label{fig:forget}
\end{figure*}

\begin{figure}[tb]
\centering
\includegraphics[width=\linewidth]{"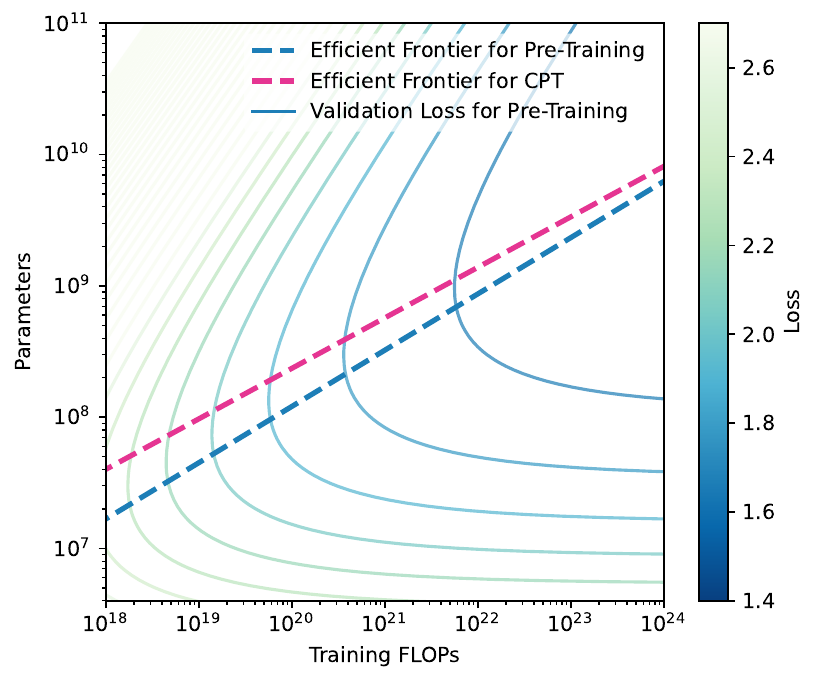"}
\caption{Predicted compute-optimal efficient frontiers on IsoLoss contour for both strategies.}
\label{fig:isoloss}
\end{figure}

When total computational resources are limited, there exists a trade-off between model parameter size and the amount of training data during pre-training.

According to the framework established in Section~\ref{sec:method}, we can determine the optimal allocation between the model parameters $N_{opt}$ and training data $D_{opt}$ by minimizing the predicted loss $L$ with respect to data $D$ and parameter size $N$. More specifically, by optimizing Equation~\ref{eq:optimal}, we estimate the optimal training data and model parameters for pre-training from scratch to be:

\begin{equation}
\begin{aligned}
    N_{opt}(C) &= 0.324 C^{0.429} \\
    D_{opt}(C) &= 0.514 C^{0.571}
\end{aligned}
\end{equation}

In comparison, for continual pre-training, the optimal allocations are:

\begin{equation}
\begin{aligned}
    \hat N_{opt}(C) = 4.79 C^{0.385}\\
    \hat D_{opt}(C) = 0.035 C^{0.615}
\end{aligned}
\end{equation}

A visualization of the efficient frontier of model parameter $N$ with respect to compute over the IsoLoss contour is shown in Figure~\ref{fig:isoloss}. We find that the optimal parameters for continual pre-training differ from those for pre-training from scratch, favoring less compute for the same model sizes. This aligns with the nature of cross-lingual transfer learning, where the model in continual pre-training is "pre-matured" due to prior knowledge acquired in the source language. This suggests that, in continual pre-training, using a larger language model is preferred over training on a larger dataset. 

It is worth noting that under our settings, larger models not only imply higher model capacity but also involve training on more data in the source language. This may explain why the compute-optimal allocation favors larger base models to some extent. However, this preference may not hold when a larger initialization model checkpoint is under-trained.

\subsection{Does Replaying from Source Language Prevent Catastrophic Forgetting?}
\label{sec:replay}

By continually pre-training a model from the source language, its performance on the target language can be greatly improved. However, with straightforward pre-training strategies, the model's performance on the source language degrades significantly. For example, in a 1.4 billion parameter model, the validation loss on English increases from 2.40 to 3.68 during pre-training. This issue is even more severe in smaller models.

To prevent catastrophic forgetting of the original distributions during continual pre-training, we investigate methods that replay data from the source language during pre-training. We use the term \textit{replaying} to refer to the practice of mixing data from the source language during continual pre-training on the target language. Previous works have shown that replaying data can help prevent catastrophic forgetting in continual learning tasks~\cite{Ibrahim2024SimpleAS, Scialom2022FinetunedLM}.

For models with 1.4B parameters, we continually train several models with mixed training corpora by replaying data at various ratios. We visualize the training curves of these English-replaying models in Figure~\ref{fig:forget}. Note that in Figure~\ref{fig:forget}, the compute is specific to each language rather than the total compute during training.

Figure~\ref{fig:forget} demonstrates that replaying data from the source language significantly alters the scaling behavior in an intricate manner. As shown on the right side of Figure~\ref{fig:forget}, different ratios of replaying only affect the early stage of training. Models reach the same validation loss when the same amount of compute is used, regardless of the varying ratios of original data, ranging from 1\% to 80\%. 

The left side of Figure~\ref{fig:forget} compares the relationship between compute and validation loss on the original distribution throughout continual pre-training, which can be viewed as the \textit{"scaling law of forgetting"}. Interestingly, the scaling behavior depicts a power-law relationship similar to that during pre-training from scratch. Validation losses of models at different English replaying ratios increase at the early stage of training and then decline, eventually returning to a lower value than at the start. This suggests that a large amount of original knowledge is preserved throughout continual training, even with a very low English replaying ratio (1\% - 5\%). Above discoveries suggest that higher levels of replaying original data are beneficial, as replaying does not hinder the scaling properties on the target language while preserving the model's performance on the original distribution.

\begin{tcolorbox}[callout={Key Takeaways}]
    \begin{enumerate}[\textbullet]
        \item Under computational constraints, a larger parameter size is preferred over pretraining on a larger dataset in CPT.
        \item Continual pre-training without replaying data from source language causes severe catastrophic forgetting, especially in smaller models. 
        \item 5\% - 30\% of source language replaying effectively prevents forgetting while not hindering efficiency of continual pre-training.
    \end{enumerate}
\end{tcolorbox}

\section{Related Work}
\label{sec:related}

\paragraph{Scaling Law for Large Language Models}
Scaling laws help us understand how model performance changes with the amount of data and the size of the model. \citet{kaplan2020scaling} first introduced a detailed scaling law for large language models, demonstrating a clear relationship between model size, training data, and performance. \citet{hoffmann2022training} further explored this by emphasizing the trade-off between the data quantity and the model size, suggesting a compute-optimal allocation of data and parameters. Recent studies have examined scaling laws under specific conditions. \citet{hernandez2022scaling} and \citet{muennighoff2023scaling} focused on the diminishing returns from repeated tokens and excessive parameters. \citet{tay2022scaling} and \citet{frantar2023scaling} investigated how different model architectures impact scaling. Scaling laws are also relevant in the context of newer pre-training methods, such as parameter-efficient fine-tuning (PEFT)~\cite{kalajdzievski2024scaling} and Mixture-of-Experts (MoE)~\cite{krajewski2024scaling}.

\paragraph{Cross-Lingual Transfer Learning}
Transfer learning aims to enhance performance on new tasks by adapting pre-trained models with out-of-domain data. This process is more efficient when the source and target domains are closely related~\cite{pan2009survey, zhuang2020comprehensive}. Cross-lingual pre-training leverages language-independent knowledge embedded in pre-trained LLMs to improve performance in the target language~\cite{wu2019emerging, yosinski2014transferable}. Transfer learning is often studied within the context of limited-scale post-training, but it has been shown to be effective at a large pre-training scale with proper techniques~\cite{gupta2023continual}. A significant challenge in transfer learning is \textit{catastrophic forgetting}~\cite{winata2023overcoming}, where the model's ability in the original training domain degrades during transfer learning. Various strategies have been proposed to mitigate catastrophic forgetting, including modified learning rate schedules~\cite{ibrahim2024simple, gupta2023continual, winata2023overcoming}, data replay~\cite{ostapenko2022continual}, and regularization~\cite{farajtabar2020orthogonal}. Our work combines data replay and modified learning rate schedules to combat catastrophic forgetting.

Our research is closely related to \citet{hernandez2021scaling}, which focused on meta-knowledge transfer between English and code under self-supervised fine-tuning settings. In contrast, we expand continual pre-training to larger-scale and cross-lingual settings, addressing the gap in effective transfer at scale for continual pre-training with significant distribution shifts.

\section{Conclusion}

In this paper, we explored continual pre-training (CPT), analyzing its principles, influencing factors, and best practices. Through training multiple LLMs with varying sizes, language distributions, and conditions, we derived an extended scaling law for CPT. Our results quantitatively demonstrate that CPT achieves lower loss more quickly, saving 25\% to 50\% of training resources. However, CPT is particularly sensitive to factors such as language type, training duration, and catastrophic forgetting. Based on these insights, we provide best practices for CPT, including optimal data-to-parameter allocation and replay ratios. These findings motivate future practitioners to apply CPT, offering deeper insights into factors like dataset distribution and training budgets. 

\section*{Limitations}

\paragraph{Language Contamination} We used public datasets for pre-training, but completely preventing English contamination is challenging, especially since languages like French often include English words. Counting samples in each language split to estimate computational effort may be imprecise if other languages are present. Future research should analyze the impact of language contamination in multilingual pre-training more deeply.

\paragraph{Hyper-Parameter Sensitivity} We selected hyper-parameters based on experience and trial and error when training models of various scales. Deviating from optimal hyper-parameters can significantly harm optimization and disrupt scaling laws. To maintain consistency, we used constant hyper-parameters matching the model scale, aligning with previous studies. Future research should find optimal hyper-parameters from the perspective of language-specific scaling laws for more effective pre-training.

\paragraph{Scaling Constraints} Due to computational limitations, we couldn't conduct extensive experiments with large datasets or very large models, which may limit the generalizability of our findings to larger-scale scenarios. We focused on the LLaMA2 architecture, known for its practicality in measuring scaling properties. However, different architectures may exhibit distinct scaling behaviors. Future research should investigate these differences to better optimize and scale various model architectures.

\paragraph{Vocabulary Extension} We did not test the impact of vocabulary extension during continual pre-training (CPT). Instead, we used a byte-level BPE tokenizer trained on both English and Chinese text, keeping the model's shape unchanged. While this allowed us to focus on scaling properties, it does not reflect practical scenarios where extending the vocabulary to include new tokens is necessary. This omission limits the applicability of our findings to cases where the tokenizer can not handle the new language well. Future work should explore vocabulary extension effects in CPT to provide a more comprehensive understanding.

\paragraph{Pre-training Length Variations} Our study assumed source models were trained to the Chinchilla-optimal token count. In practice, models are often "over-trained" beyond this point, such as LLaMA2 and LLaMA3 trained with trillions of tokens. We did not investigate how continual pre-training scales for these over-trained models. This limits the applicability of our scaling laws to real-world scenarios where models exceed the Chinchilla-optimal training length. Future research should examine CPT scaling for over-trained models to determine if our conclusions hold in such settings.

\section*{Acknowledgements}
We would like to extend our special thanks to Yadong Liu and Chunhui Liu for their invaluable feedback and support.

\bibliography{anthology,custom}

\begin{thebibliography}{43}
\expandafter\ifx\csname natexlab\endcsname\relax\def\natexlab#1{#1}\fi

\bibitem[{Achiam et~al.(2023)Achiam, Adler, Agarwal, Ahmad, Akkaya, Aleman, Almeida, Altenschmidt, Altman, Anadkat et~al.}]{achiam2023gpt}
Josh Achiam, Steven Adler, Sandhini Agarwal, Lama Ahmad, Ilge Akkaya, Florencia~Leoni Aleman, Diogo Almeida, Janko Altenschmidt, Sam Altman, Shyamal Anadkat, et~al. 2023.
\newblock Gpt-4 technical report.
\newblock \emph{arXiv preprint arXiv:2303.08774}.

\bibitem[{Bisk et~al.(2019)Bisk, Zellers, Bras, Gao, and Choi}]{Bisk2019PIQARA}
Yonatan Bisk, Rowan Zellers, Ronan~Le Bras, Jianfeng Gao, and Yejin Choi. 2019.
\newblock \href {https://api.semanticscholar.org/CorpusID:208290939} {Piqa: Reasoning about physical commonsense in natural language}.
\newblock \emph{ArXiv}, abs/1911.11641.

\bibitem[{Brown et~al.(2020)Brown, Mann, Ryder, Subbiah, Kaplan, Dhariwal, Neelakantan, Shyam, Sastry, Askell et~al.}]{brown2020language}
Tom Brown, Benjamin Mann, Nick Ryder, Melanie Subbiah, Jared~D Kaplan, Prafulla Dhariwal, Arvind Neelakantan, Pranav Shyam, Girish Sastry, Amanda Askell, et~al. 2020.
\newblock Language models are few-shot learners.
\newblock \emph{Advances in neural information processing systems}, 33:1877--1901.

\bibitem[{{Common Crawl}(2007)}]{commoncrawl}
{Common Crawl}. 2007.
\newblock Common crawl: A public repository of web crawl data.
\newblock \url{https://commoncrawl.org/}.
\newblock Accessed: 2024-09-26.

\bibitem[{Computer(2023)}]{together2023redpajama}
Together Computer. 2023.
\newblock \href {https://github.com/togethercomputer/RedPajama-Data} {Redpajama: an open dataset for training large language models}.

\bibitem[{Conneau et~al.(2018)Conneau, Rinott, Lample, Williams, Bowman, Schwenk, and Stoyanov}]{conneau2018xnli}
Alexis Conneau, Ruty Rinott, Guillaume Lample, Adina Williams, Samuel~R. Bowman, Holger Schwenk, and Veselin Stoyanov. 2018.
\newblock Xnli: Evaluating cross-lingual sentence representations.
\newblock In \emph{Proceedings of the 2018 Conference on Empirical Methods in Natural Language Processing}. Association for Computational Linguistics.

\bibitem[{Dac~Lai et~al.(2023)Dac~Lai, Van~Nguyen, Ngo, Nguyen, Dernoncourt, Rossi, and Nguyen}]{dac2023okapi}
Viet Dac~Lai, Chien Van~Nguyen, Nghia~Trung Ngo, Thuat Nguyen, Franck Dernoncourt, Ryan~A Rossi, and Thien~Huu Nguyen. 2023.
\newblock Okapi: Instruction-tuned large language models in multiple languages with reinforcement learning from human feedback.
\newblock \emph{arXiv e-prints}, pages arXiv--2307.

\bibitem[{Eronen et~al.(2023)Eronen, Ptaszynski, and Masui}]{eronen2023zero}
Juuso Eronen, Michal Ptaszynski, and Fumito Masui. 2023.
\newblock Zero-shot cross-lingual transfer language selection using linguistic similarity.
\newblock \emph{Information Processing \& Management}, 60(3):103250.

\bibitem[{Farajtabar et~al.(2020)Farajtabar, Azizan, Mott, and Li}]{farajtabar2020orthogonal}
Mehrdad Farajtabar, Navid Azizan, Alex Mott, and Ang Li. 2020.
\newblock Orthogonal gradient descent for continual learning.
\newblock In \emph{International Conference on Artificial Intelligence and Statistics}, pages 3762--3773. PMLR.

\bibitem[{Frantar et~al.(2023)Frantar, Riquelme, Houlsby, Alistarh, and Evci}]{frantar2023scaling}
Elias Frantar, Carlos Riquelme, Neil Houlsby, Dan Alistarh, and Utku Evci. 2023.
\newblock Scaling laws for sparsely-connected foundation models.
\newblock \emph{arXiv preprint arXiv:2309.08520}.

\bibitem[{Gupta et~al.(2023)Gupta, Th{\'e}rien, Ibrahim, Richter, Anthony, Belilovsky, Rish, and Lesort}]{gupta2023continual}
Kshitij Gupta, Benjamin Th{\'e}rien, Adam Ibrahim, Mats~L Richter, Quentin Anthony, Eugene Belilovsky, Irina Rish, and Timoth{\'e}e Lesort. 2023.
\newblock Continual pre-training of large language models: How to (re) warm your model?
\newblock \emph{arXiv preprint arXiv:2308.04014}.

\bibitem[{Hernandez et~al.(2022)Hernandez, Brown, Conerly, DasSarma, Drain, El-Showk, Elhage, Hatfield-Dodds, Henighan, Hume et~al.}]{hernandez2022scaling}
Danny Hernandez, Tom Brown, Tom Conerly, Nova DasSarma, Dawn Drain, Sheer El-Showk, Nelson Elhage, Zac Hatfield-Dodds, Tom Henighan, Tristan Hume, et~al. 2022.
\newblock Scaling laws and interpretability of learning from repeated data.
\newblock \emph{arXiv preprint arXiv:2205.10487}.

\bibitem[{Hernandez et~al.(2021)Hernandez, Kaplan, Henighan, and McCandlish}]{hernandez2021scaling}
Danny Hernandez, Jared Kaplan, Tom Henighan, and Sam McCandlish. 2021.
\newblock Scaling laws for transfer.
\newblock \emph{arXiv preprint arXiv:2102.01293}.

\bibitem[{Hoffmann et~al.(2022)Hoffmann, Borgeaud, Mensch, Buchatskaya, Cai, Rutherford, Casas, Hendricks, Welbl, Clark et~al.}]{hoffmann2022training}
Jordan Hoffmann, Sebastian Borgeaud, Arthur Mensch, Elena Buchatskaya, Trevor Cai, Eliza Rutherford, Diego de~Las Casas, Lisa~Anne Hendricks, Johannes Welbl, Aidan Clark, et~al. 2022.
\newblock Training compute-optimal large language models.
\newblock \emph{arXiv preprint arXiv:2203.15556}.

\bibitem[{Huber(1992)}]{huber1992robust}
Peter~J Huber. 1992.
\newblock Robust estimation of a location parameter.
\newblock In \emph{Breakthroughs in statistics: Methodology and distribution}, pages 492--518. Springer.

\bibitem[{Ibrahim et~al.(2024{\natexlab{a}})Ibrahim, Th'erien, Gupta, Richter, Anthony, Lesort, Belilovsky, and Rish}]{Ibrahim2024SimpleAS}
Adam Ibrahim, Benjamin Th'erien, Kshitij Gupta, Mats~L. Richter, Quentin Anthony, Timoth{\'e}e Lesort, Eugene Belilovsky, and Irina Rish. 2024{\natexlab{a}}.
\newblock \href {https://api.semanticscholar.org/CorpusID:268379604} {Simple and scalable strategies to continually pre-train large language models}.
\newblock \emph{ArXiv}, abs/2403.08763.

\bibitem[{Ibrahim et~al.(2024{\natexlab{b}})Ibrahim, Th{\'e}rien, Gupta, Richter, Anthony, Lesort, Belilovsky, and Rish}]{ibrahim2024simple}
Adam Ibrahim, Benjamin Th{\'e}rien, Kshitij Gupta, Mats~L Richter, Quentin Anthony, Timoth{\'e}e Lesort, Eugene Belilovsky, and Irina Rish. 2024{\natexlab{b}}.
\newblock Simple and scalable strategies to continually pre-train large language models.
\newblock \emph{arXiv preprint arXiv:2403.08763}.

\bibitem[{Kalajdzievski(2024)}]{kalajdzievski2024scaling}
Damjan Kalajdzievski. 2024.
\newblock Scaling laws for forgetting when fine-tuning large language models.
\newblock \emph{arXiv preprint arXiv:2401.05605}.

\bibitem[{Kaplan et~al.(2020)Kaplan, McCandlish, Henighan, Brown, Chess, Child, Gray, Radford, Wu, and Amodei}]{kaplan2020scaling}
Jared Kaplan, Sam McCandlish, Tom Henighan, Tom~B Brown, Benjamin Chess, Rewon Child, Scott Gray, Alec Radford, Jeffrey Wu, and Dario Amodei. 2020.
\newblock Scaling laws for neural language models.
\newblock \emph{arXiv preprint arXiv:2001.08361}.

\bibitem[{Krajewski et~al.(2024)Krajewski, Ludziejewski, Adamczewski, Pi{\'o}ro, Krutul, Antoniak, Ciebiera, Kr{\'o}l, Odrzyg{\'o}{\'z}d{\'z}, Sankowski et~al.}]{krajewski2024scaling}
Jakub Krajewski, Jan Ludziejewski, Kamil Adamczewski, Maciej Pi{\'o}ro, Micha{\l} Krutul, Szymon Antoniak, Kamil Ciebiera, Krystian Kr{\'o}l, Tomasz Odrzyg{\'o}{\'z}d{\'z}, Piotr Sankowski, et~al. 2024.
\newblock Scaling laws for fine-grained mixture of experts.
\newblock \emph{arXiv preprint arXiv:2402.07871}.

\bibitem[{Kudo(2018)}]{kudo2018sentencepiece}
T~Kudo. 2018.
\newblock Sentencepiece: A simple and language independent subword tokenizer and detokenizer for neural text processing.
\newblock \emph{arXiv preprint arXiv:1808.06226}.

\bibitem[{Lin et~al.(2021)Lin, Mihaylov, Artetxe, Wang, Chen, Simig, Ott, Goyal, Bhosale, Du, Pasunuru, Shleifer, Koura, Chaudhary, O'Horo, Wang, Zettlemoyer, Kozareva, Diab, Stoyanov, and Li}]{Lin2021FewshotLWStorycloze}
Xi~Victoria Lin, Todor Mihaylov, Mikel Artetxe, Tianlu Wang, Shuohui Chen, Daniel Simig, Myle Ott, Naman Goyal, Shruti Bhosale, Jingfei Du, Ramakanth Pasunuru, Sam Shleifer, Punit~Singh Koura, Vishrav Chaudhary, Brian O'Horo, Jeff Wang, Luke Zettlemoyer, Zornitsa Kozareva, Mona~T. Diab, Ves Stoyanov, and Xian Li. 2021.
\newblock \href {https://api.semanticscholar.org/CorpusID:245334784} {Few-shot learning with multilingual generative language models}.
\newblock In \emph{Conference on Empirical Methods in Natural Language Processing}.

\bibitem[{Muennighoff et~al.(2023)Muennighoff, Rush, Barak, Scao, Piktus, Tazi, Pyysalo, Wolf, and Raffel}]{muennighoff2023scaling}
Niklas Muennighoff, Alexander~M Rush, Boaz Barak, Teven~Le Scao, Aleksandra Piktus, Nouamane Tazi, Sampo Pyysalo, Thomas Wolf, and Colin Raffel. 2023.
\newblock Scaling data-constrained language models.
\newblock \emph{arXiv preprint arXiv:2305.16264}.

\bibitem[{Nocedal(1980)}]{nocedal1980updating}
Jorge Nocedal. 1980.
\newblock Updating quasi-newton matrices with limited storage.
\newblock \emph{Mathematics of computation}, 35(151):773--782.

\bibitem[{Ostapenko et~al.(2022)Ostapenko, Lesort, Rodr{\'\i}guez, Arefin, Douillard, Rish, and Charlin}]{ostapenko2022continual}
Oleksiy Ostapenko, Timothee Lesort, Pau Rodr{\'\i}guez, Md~Rifat Arefin, Arthur Douillard, Irina Rish, and Laurent Charlin. 2022.
\newblock Continual learning with foundation models: An empirical study of latent replay.
\newblock In \emph{Conference on lifelong learning agents}, pages 60--91. PMLR.

\bibitem[{Pan and Yang(2009)}]{pan2009survey}
Sinno~Jialin Pan and Qiang Yang. 2009.
\newblock A survey on transfer learning.
\newblock \emph{IEEE Transactions on knowledge and data engineering}, 22(10):1345--1359.

\bibitem[{Ponti et~al.(2020)Ponti, Glavavs, Majewska, Liu, Vulic, and Korhonen}]{Ponti2020XCOPAAM}
E.~Ponti, Goran Glavavs, Olga Majewska, Qianchu Liu, Ivan Vulic, and Anna Korhonen. 2020.
\newblock \href {https://api.semanticscholar.org/CorpusID:218470125} {Xcopa: A multilingual dataset for causal commonsense reasoning}.
\newblock In \emph{Conference on Empirical Methods in Natural Language Processing}.

\bibitem[{Radford et~al.(2019)Radford, Wu, Child, Luan, Amodei, Sutskever et~al.}]{radford2019language}
Alec Radford, Jeffrey Wu, Rewon Child, David Luan, Dario Amodei, Ilya Sutskever, et~al. 2019.
\newblock Language models are unsupervised multitask learners.
\newblock \emph{OpenAI blog}, 1(8):9.

\bibitem[{Raffel et~al.(2019)Raffel, Shazeer, Roberts, Lee, Narang, Matena, Zhou, Li, and Liu}]{2019t5}
Colin Raffel, Noam Shazeer, Adam Roberts, Katherine Lee, Sharan Narang, Michael Matena, Yanqi Zhou, Wei Li, and Peter~J. Liu. 2019.
\newblock \href {http://arxiv.org/abs/1910.10683} {Exploring the limits of transfer learning with a unified text-to-text transformer}.
\newblock \emph{arXiv e-prints}.

\bibitem[{Sakaguchi et~al.(2019)Sakaguchi, Bras, Bhagavatula, and Choi}]{Sakaguchi2019WinoGrande}
Keisuke Sakaguchi, Ronan~Le Bras, Chandra Bhagavatula, and Yejin Choi. 2019.
\newblock \href {https://api.semanticscholar.org/CorpusID:198893658} {Winogrande}.
\newblock \emph{Communications of the ACM}, 64:99 -- 106.

\bibitem[{Scialom et~al.(2022)Scialom, Chakrabarty, and Muresan}]{Scialom2022FinetunedLM}
Thomas Scialom, Tuhin Chakrabarty, and Smaranda Muresan. 2022.
\newblock \href {https://api.semanticscholar.org/CorpusID:252815378} {Fine-tuned language models are continual learners}.
\newblock In \emph{Conference on Empirical Methods in Natural Language Processing}.

\bibitem[{Siegel and Xu(2020)}]{siegel2020approximation}
Jonathan~W Siegel and Jinchao Xu. 2020.
\newblock Approximation rates for neural networks with general activation functions.
\newblock \emph{Neural Networks}, 128:313--321.

\bibitem[{Tan et~al.(2018)Tan, Sun, Kong, Zhang, Yang, and Liu}]{tan2018survey}
Chuanqi Tan, Fuchun Sun, Tao Kong, Wenchang Zhang, Chao Yang, and Chunfang Liu. 2018.
\newblock A survey on deep transfer learning.
\newblock In \emph{Artificial Neural Networks and Machine Learning--ICANN 2018: 27th International Conference on Artificial Neural Networks, Rhodes, Greece, October 4-7, 2018, Proceedings, Part III 27}, pages 270--279. Springer.

\bibitem[{Tang et~al.(2020)Tang, Tran, Li, Chen, Goyal, Chaudhary, Gu, and Fan}]{tang2020multilingual}
Yuqing Tang, Chau Tran, Xian Li, Peng-Jen Chen, Naman Goyal, Vishrav Chaudhary, Jiatao Gu, and Angela Fan. 2020.
\newblock Multilingual translation with extensible multilingual pretraining and finetuning.
\newblock \emph{arXiv preprint arXiv:2008.00401}.

\bibitem[{Tay et~al.(2022)Tay, Dehghani, Abnar, Chung, Fedus, Rao, Narang, Tran, Yogatama, and Metzler}]{tay2022scaling}
Yi~Tay, Mostafa Dehghani, Samira Abnar, Hyung~Won Chung, William Fedus, Jinfeng Rao, Sharan Narang, Vinh~Q Tran, Dani Yogatama, and Donald Metzler. 2022.
\newblock Scaling laws vs model architectures: How does inductive bias influence scaling?
\newblock \emph{arXiv preprint arXiv:2207.10551}.

\bibitem[{Touvron et~al.(2023)Touvron, Martin, Stone, Albert, Almahairi, Babaei, Bashlykov, Batra, Bhargava, Bhosale et~al.}]{touvron2023llama2}
Hugo Touvron, Louis Martin, Kevin Stone, Peter Albert, Amjad Almahairi, Yasmine Babaei, Nikolay Bashlykov, Soumya Batra, Prajjwal Bhargava, Shruti Bhosale, et~al. 2023.
\newblock Llama 2: Open foundation and fine-tuned chat models.
\newblock \emph{arXiv preprint arXiv:2307.09288}.

\bibitem[{Wei et~al.(2022)Wei, Tay, Bommasani, Raffel, Zoph, Borgeaud, Yogatama, Bosma, Zhou, Metzler et~al.}]{wei2022emergent}
Jason Wei, Yi~Tay, Rishi Bommasani, Colin Raffel, Barret Zoph, Sebastian Borgeaud, Dani Yogatama, Maarten Bosma, Denny Zhou, Donald Metzler, et~al. 2022.
\newblock Emergent abilities of large language models.
\newblock \emph{arXiv preprint arXiv:2206.07682}.

\bibitem[{Winata et~al.(2023)Winata, Xie, Radhakrishnan, Wu, Jin, Cheng, Kulkarni, and Preotiuc-Pietro}]{winata2023overcoming}
Genta~Indra Winata, Lingjue Xie, Karthik Radhakrishnan, Shijie Wu, Xisen Jin, Pengxiang Cheng, Mayank Kulkarni, and Daniel Preotiuc-Pietro. 2023.
\newblock Overcoming catastrophic forgetting in massively multilingual continual learning.
\newblock \emph{arXiv preprint arXiv:2305.16252}.

\bibitem[{Wu et~al.(2019)Wu, Conneau, Li, Zettlemoyer, and Stoyanov}]{wu2019emerging}
Shijie Wu, Alexis Conneau, Haoran Li, Luke Zettlemoyer, and Veselin Stoyanov. 2019.
\newblock Emerging cross-lingual structure in pretrained language models.
\newblock \emph{arXiv preprint arXiv:1911.01464}.

\bibitem[{Xue et~al.(2022)Xue, Zhao, Yuan, and Wang}]{wudao2022}
Zhao Xue, Hanyu Zhao, Sha Yuan, and Yequan Wang. 2022.
\newblock \href {https://doi.org/10.57760/sciencedb.o00126.00004} {{WuDaoCorpora Text}}.

\bibitem[{Yosinski et~al.(2014)Yosinski, Clune, Bengio, and Lipson}]{yosinski2014transferable}
Jason Yosinski, Jeff Clune, Yoshua Bengio, and Hod Lipson. 2014.
\newblock How transferable are features in deep neural networks?
\newblock \emph{Advances in neural information processing systems}, 27.

\bibitem[{Zhang et~al.(2024)Zhang, Liu, Cherry, and Firat}]{zhang2024scaling}
Biao Zhang, Zhongtao Liu, Colin Cherry, and Orhan Firat. 2024.
\newblock When scaling meets llm finetuning: The effect of data, model and finetuning method.
\newblock \emph{arXiv preprint arXiv:2402.17193}.

\bibitem[{Zhuang et~al.(2020)Zhuang, Qi, Duan, Xi, Zhu, Zhu, Xiong, and He}]{zhuang2020comprehensive}
Fuzhen Zhuang, Zhiyuan Qi, Keyu Duan, Dongbo Xi, Yongchun Zhu, Hengshu Zhu, Hui Xiong, and Qing He. 2020.
\newblock A comprehensive survey on transfer learning.
\newblock \emph{Proceedings of the IEEE}, 109(1):43--76.

\end{thebibliography}
\bibliographystyle{acl_natbib}

\appendix
\section{Downstream Performance of English-Replaying Models at Various Ratios}

\begin{figure}[htbp]
  \centering
  \includegraphics[width=\linewidth]{"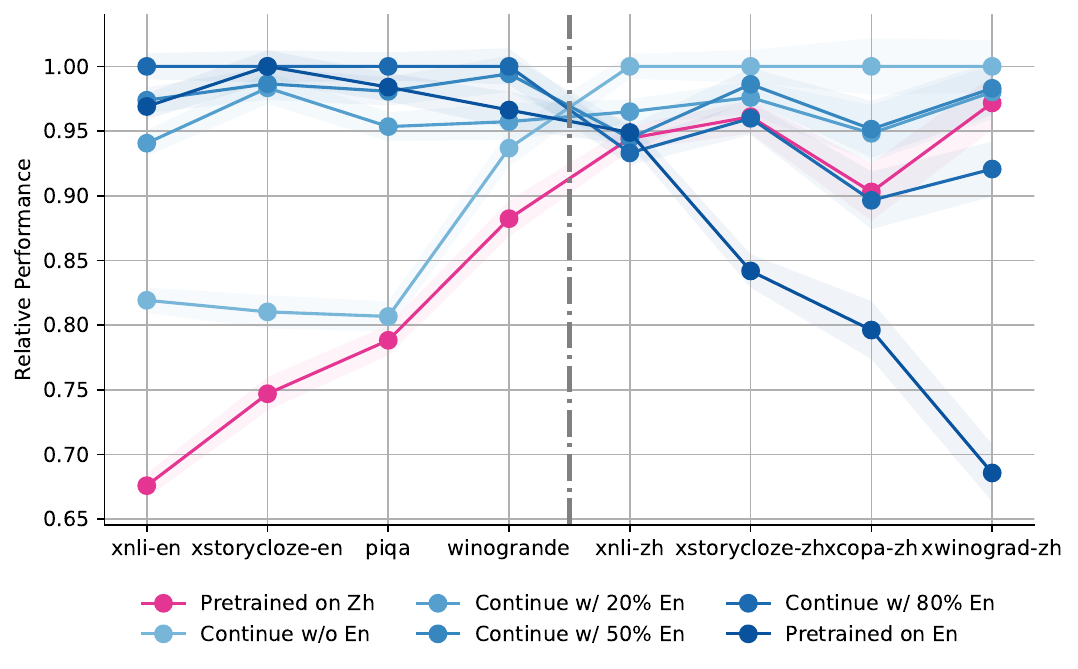"}
  \caption{Model performance on English and Chinese benchmarks at different English data replaying ratios with 1.4B parameters. Relative Performance refers to accuracy relative to the highest accuracy achieved across different training settings with 1.4B parameters.}
  \label{fig:en_ratio}
  \end{figure}

To further analyze the impacts of mixing original data in continual pre-training, we evaluate model performance on English and Chinese benchmarks at different English data mix ratios in Figure~\ref{fig:en_ratio}. The results show that pre-training solely on one language leads to sub-optimal performance on the other language. However, incorporating even a small amount of English data can effectively maintain performance on both original distributions. In practice, around 30\% of original data is sufficient to keep the validation loss lower than at the start of continual pre-training.

Models pre-trained only on English excel on English benchmarks but perform poorly on Chinese benchmarks, and vice versa. Adding English data to models initially pre-trained on Chinese improves their English performance without significantly harming their Chinese performance. This improvement is observed across different proportions of English data (20\%, 50\%, and 80\%).An optimal ratio is around 30\% English data, balancing low validation loss and high relative performance across both languages. Beyond 50\% English data, there are diminishing returns, with marginal gains in English performance and a slight decline in Chinese performance. 

\section{Fitting Error for Extended Scaling Law}
\label{sec:fiterror}
\begin{table}[h]
\centering
\caption{Comparison of fitting errors $L$ for the Chinchilla Law~\cite{hoffmann2022training} and our extended scaling law on empirical data. The fitt error in huber loss is denoted as $L_{equation}$. Our extended scaling law performs better for CPT, comparable to Chinchilla in pre-training.}
\begin{tabularx}{\linewidth}{XXX}
\toprule
Fit Data & Pre-Training & CPT \\
\midrule
$L_{Chinchilla}$ & 0.0090 & 0.0108 \\
$L_{Ours}$ & 0.0094 & 0.0093 \\
\midrule
$\gamma$ in Eq.~\ref{eq:oursloss} & -0.005 & 0.080 \\
\bottomrule
\end{tabularx}
\label{tab:fiterror}
\end{table}

We applied the Chinchilla Law~\cite{hoffmann2022training} and our extended scaling law to empirical data from both pre-training from scratch and continual pre-training (CPT) on Chinese text. The fitting process minimized the average loss across all trained models for both strategies using the same procedures described in Section~\ref{sec:fit}. The results, shown in Table~\ref{tab:fiterror}, indicate that for pre-training from scratch, the extended scaling law performs similarly to the Chinchilla Law, with the factor $\gamma$ close to zero. In contrast, for continual pre-training, the joint data-parameter term in the extended scaling law significantly reduces the fitting error, with $\gamma = 0.080$.

\section{Theoretical Analysis and Interpretation of Extended Scaling Law}
\label{app:explan}
First, we review the formulated scaling law proposed by~\citet{hoffmann2022training}, where they derived and fit a formula for the loss. They decompose the loss $L(N, D)$ into three terms in the abstract functional space:
\begin{equation}
\begin{aligned}
L(N, D) \triangleq & \ L(\bar{f}_{N,D}) \\
= & \ L(f^*) + \left(L(\hat{f}_{N}) - L(f^*)\right) \\
& + \left(L(\bar{f}_{N,D}) - L(\hat{f}_{N})\right)
\label{eq:chinchilla_1}
\end{aligned}
\end{equation}

Here, $N$ represents the parameters, $D$ represents the training tokens, $f^*$ represents the optimal Bayesian classifier, $\hat{f}_{N}$ denotes the optimal transformer model under the constraint of parameters $N$, $\bar{f}_{N,D}$ represents the outcome obtained through gradient descent under the constraints of parameters $N$ and training tokens $D$ in the experiments. 

This functional space decomposition includes three parts:the Bayes risk $L(f^*)$, which is the smallest possible loss for predicting the next token based on the full distribution $P$, also known as the "entropy of natural text", a term $\left(L(\hat{f}_{N}) - L(f^*)\right)$ related to how well the function approximates based on the hypothesis space size, and a stochastic approximation term $\left(L(\bar{f}_{N,D}) - L(\hat{f}_{N})\right)$.

\paragraph{Functional space decomposition}

Our goal is to modify the Equation~\ref{eq:chinchilla_2} to fit the scenario of continual pre-training. Consider Continual Pre-training as initialization from a specific model weight state, recalling the functional space decomposition -- Equation~\ref{eq:chinchilla_1}. It serves as a loss decomposition under token and model size constraints, discuss in the abstract functional space. This decomposition method has no relation to the training process (including initialization, naturally), but is a theoretical analysis and summary, so we think that the structure of the entire decomposition is unaffected. 

Keeping the structure of Equation~\ref{eq:chinchilla_1}, let's continue to analyze the impact on the each three term. When considering continual pre-training as a form of random initialization, recall the meaning of the first two terms: the entropy of natural text and the restrictions on the scale of the parameter space, they are both independent of the specific training process and only depend on the model's architecture, as well as the scale of $N$ and $D$. Therefore, different initialization will only affect The process we implement gradient descent, which is the last term: $L(\bar{f}_{N,D}) - L(\hat{f}_{N})$.

Overall, in this scenario, we inherit Equation~\ref{eq:chinchilla_1} and then fine-tuned Equation~\ref{eq:chinchilla_2}.

\paragraph{Inheriting learned variables}

Pay attention to the detailed settings of our training scenario. the dataset used for training and the details of the entire training process are consistent. We will discuss the expected forms and explain the reasons for inheriting learned variables:

(1) For the first term, $L(f^*)$, due to the consistency of the dataset, the entropy of training data naturally maintain consistency between continual pre-training and training from scratch. Numerically, this is equivalent to the same constant E.

(2) For the second term, $L(\hat{f}_{N}) - L(f^*)$, depends entirely on the number of parameters N that defines the size of the functional approximation space. Siegel and Xu (2020)\cite{siegel2020approximation} analyzed this term and found it is related to the power of N. We inherit this perspective and believe that its estimated form is $\frac{A}{N^\alpha}$. From the principle of decomposition, this second term does not involve the training phase and only represents the abstract restriction of model's parameter scale. When comparing to training from scratch, the model’s size N and architecture are completely consistent, so we inherits the values of $A$ and $\alpha$. 

\onecolumn{
\begin{center}
\section{Model Structural Parameters}
\end{center}
\begin{table*}[htbp]
  \centering
  \small
  \caption{Structural Parameters for Models of Different Sizes.}
  \begin{tabularx}{\textwidth}{XXXXX}
  \toprule
  Parameter Size(M) & Hidden Layer Size & Intermediate Layer & Attention Head Count & Number of Layers \\
  \midrule
  49  & 512   & 3072  & 8  & 8  \\
  66  & 576   & 3584  & 9  & 9  \\
  86  & 640   & 3584  & 10 & 10 \\
  105 & 640   & 3584  & 10 & 13 \\
  125 & 640   & 3584  & 10 & 16 \\
  137 & 768   & 4608  & 12 & 12 \\
  166 & 768   & 4608  & 12 & 15 \\
  194 & 768   & 4608  & 12 & 18 \\
  208 & 896   & 5120  & 14 & 14 \\
  234 & 896   & 5120  & 14 & 16 \\
  259 & 896   & 5120  & 14 & 18 \\
  301 & 1024  & 5632  & 16 & 16 \\
  334 & 1024  & 5632  & 16 & 18 \\
  368 & 1024  & 5632  & 16 & 20 \\
  512 & 1280  & 7168  & 10 & 18 \\
  591 & 1280  & 7168  & 10 & 21 \\
  616 & 1408  & 7680  & 11 & 18 \\
  670  & 1280  & 7168  & 10 & 24 \\
  711  & 1408  & 7680  & 11 & 21 \\
  766  & 1536  & 8704  & 12 & 19 \\
  806  & 1408  & 7680  & 11 & 24 \\
  879  & 1536  & 8704  & 12 & 22 \\
  992  & 1536  & 8704  & 12 & 25 \\
  1085 & 1792  & 9728  & 14 & 20 \\
  1239 & 1792  & 9728  & 14 & 23 \\
  1393 & 1792  & 9728  & 14 & 26 \\
  1542 & 2048  & 11264 & 16 & 22 \\
  1736 & 2176  & 11776 & 17 & 22 \\
  1743 & 2048  & 11264 & 16 & 25 \\
  1944 & 2048  & 11264 & 16 & 28 \\
  1963 & 2176  & 11776 & 17 & 25 \\
  2112 & 2304  & 12800 & 18 & 24 \\
  2191 & 2176  & 11776 & 17 & 28 \\
  2452 & 2304  & 12800 & 18 & 28 \\
  2791 & 2304  & 12800 & 18 & 32 \\
  2808 & 2560  & 13824 & 20 & 26 \\
  3227 & 2560  & 13824 & 20 & 30 \\
  3647 & 2560  & 13824 & 20 & 34 \\
  4016 & 2688  & 14848 & 22 & 34 \\
  4248 & 2688  & 14848 & 21 & 36 \\
  4657 & 2816  & 15360 & 22 & 36 \\
  5534 & 3072  & 16896 & 24 & 36 \\
  \bottomrule
  \end{tabularx}
  \label{tab:model_configs}
\end{table*}
} 
\label{app:struct}



\end{document}